\documentclass[11pt,a4paper]{article}
\usepackage[hyperref]{acl2018}
\usepackage{times}
\usepackage{latexsym}
\usepackage{subcaption}
\usepackage{color}
\usepackage{amsmath,amsfonts,amssymb}
\usepackage{footnote}

\usepackage{url}

\aclfinalcopy 

\title{Data-Driven Methods for Solving Algebra Word Problems}

\author{Benjamin Robaidek \\
  University of Washington \\
  {\tt benro@uw.edu} \\\And
  Rik Koncel-Kedziorski \\
  University of Washington \\
  {\tt kedzior@uw.edu} \\\And
  Hannaneh Hajishirzi \\
  University of Washington\\
  {\tt hannaneh@cs.washington.edu}}

\date{}

\begin{document}
\maketitle
\begin{abstract}
 We explore 
 contemporary, data-driven techniques for solving math word problems over recent large-scale datasets.
 We show that well-tuned neural equation classifiers can outperform more sophisticated models such as sequence to sequence and self-attention across these datasets. 
 Our error analysis indicates that, while fully data driven models show some promise, semantic and world knowledge is necessary for further advances. 
 
\end{abstract}

\section{Introduction}
  Solving math word problems has been an interest of the natural language processing community since the 1960s \cite{feigenbaum1963computers, bobrow1964natural}. 
  More recently, algorithms for learning to solve algebra problems have gone in complementary directions: semantic and purely data-driven. 
  
  {\it Semantic} methods learn from data how to map problem texts to a semantic representation which can then be converted to an equation. 
  These representations combine set-like constructs \cite{hosseini2014learning} with hierarchical representations like equation trees \cite{koncel2015parsing,roy2015solving,wang2018mathdqn}. 
  Such methods have the benefit of being interpretable, but no semantic representation general enough to solve all varieties of math word problems, including proportion problems and those that map to systems of equations, has been found. 
  
  Another popular line of research is on purely {\it data-driven} solvers. 
  Given enough training data, data-driven models can learn to map word problem texts to arbitrarily complex equations or systems of equations. 
  These models have the additional advantage of being more language-independent than semantic methods, which often rely on parsers and other NLP tools.
  To train these fully data driven models, large-scale datasets for both English and Chinese were recently introduced \cite{wang2017deep,koncel2016mawps}. 
  
  In response to the success of representation learning elsewhere in NLP, sequence to sequence (seq2seq) models have been applied to algebra problem solving \cite{wang2017deep}.
  These powerful models have been shown to outperform other data-driven approaches in a variety of tasks. 
  However, it is not obvious that solving word problems is best modeled as a sequence prediction task rather than a classification or retrieval task.
  Downstream applications such as question answering or automated tutoring systems may never have to deal with arbitrarily complex or even unseen equation types, obviating the need for a sequence prediction model.
  
  These considerations beg the questions: how do data-driven approaches to math word problem solving compare to each other? 
  How can data-driven approaches benefit from recent advances in neural representation learning?  
  What are the limits of data-driven solvers?
  
  In this paper, we thoroughly examine data-driven techniques on three larger algebra word problem datasets~\cite{huang2016well,koncel2016mawps,wang2017deep}. 
  We study classification, generation, and information retrieval models,  
  and examine popular extensions to these models such as structured self-attention~\cite{lin2017structured} and the use of pretrained word embeddings~\cite{pennington2014glove, peters2018deep}. 
  
  Our experiments show that a well-tuned neural equation classifier consistently performs better than more sophisticated solvers. 
  We provide evidence that pretrained word embeddings, useful in other tasks, are not helpful for word problem solving. 
  Advanced modeling such as structured self-attention is not shown to improve performance versus a well-tuned BiLSTM Classifier. 
  Our error analysis supports the idea that, while data-driven techniques are powerful and robust, many word problems require semantic or world knowledge that cannot be easily incorporated into an end-to-end learning framework.

  \section{Problem Formulation} Solving an algebra word problem (as shown below) requires finding the correct solution given the text of the problem. 
  \begin{center}
      \begin{tabular}{p{6.5cm}} 
      {\bf Problem Text} \\
       {\it  Aliyah had some candy to give to her 3 children. She first took 2 pieces for herself and then evenly divided the rest among her children. Each child received 5 pieces. With how many pieces did she start?}\\
       {\bf Equation} \\
       {\tt 2 + (3 * 5) = x} \\
       {\bf Template} \\
       {\tt B + (A * C) = x} \\
      \end{tabular}
      \end{center}

Similar to previous data-driven methods, we frame the task as one of mapping the word problem texts to equations given the training data. 
Our models abstract the specific numbers away from both the word problem text and target equation, preserving the ordering of the numbers found in the problem text.
  The resulting abstracted equation is called an {\it equation template}. 
  At inference time, our solvers produce an equation template given the test problem. 
  The template is then populated with the actual numbers from the problem text and evaluated to produce a solution.
  
  
  \section{Models}

  \subsection{Retrieval}
    Retrieval methods map test word problem texts at inference time to the nearest training problem according to some similarity metric.
    The nearest neighbor's equation template is then filled in with numbers from the test problem and solved. 
    Following~\citet{wang2017deep}, we use Jaccard distance in this model. 
    For test problem $S$ and training problem $T$, the Jaccard similarity is computed as: $\textrm{jacc}(S,T) = \frac{S\cap T}{S\cup T}$.
    We also evaluate the use of a cosine similarity metric. 
      Words from $S$ and $T$ are associated with pretrained vectors $v(w_i)$ \cite{pennington2014glove}.
      These vectors are averaged across each problem, resulting in vectors $\bf{S}$ and $\bf{T}$. The Cosine similarity is then computed as  $\textrm{cos}(\bf{S},\bf{T})=\frac{\bf{S}\cdot \bf{T}}{||\bf{S}\bf{T}||}$.
      Vector averaging has previously been used as a strong baseline for a variety of sentence similarity tasks \cite{mu2017representing}.

  \subsection{Classification}
   Classification methods learn to map problem texts to equation templates by learning parameters that minimize a cross entropy loss function over the set of training instances.
    At inference time, these methods choose the most likely equation template (the class) given a test word problem text.
    In both retrieval and classification methods, model accuracy is upper bounded by the oracle accuracy, or the number of test equation templates which appear in the training data.
    \paragraph{BiLSTM}
    The BiLSTM classification model encodes the word problem text using a bidirectional Long Short Term Memory network \cite{hochreiter1997long} with learned parameters $\theta$.
    The final hidden state of this encoding ${\bf h_n}$ is scaled to the number of classes by weights ${\bf W} = {\bf w_1...w_k}$ and passed through a softmax to produce a distribution over class labels.
    The probability of equation template $j$ for problem $S$ is given by:
    \begin{align*}
        p(\textrm{y=}j|S,{\bf W},\theta) = \frac{\exp^{\bf{{h_n}^\intercal w_j}}}{\sum_{k=1}^{K}\exp^{\bf{h_n^\intercal w_k}}}
    \end{align*}
    This model is trained end-to-end using cross entropy loss.

    \paragraph{Structured Self-Attention}
    Sentence embeddings using self-attention mechanisms \cite{lin2017structured} were shown to be  successful in question answering tasks \cite{liu2017stochastic}. 
    We conjecture that algebra problem solvers can also benefit   from the long  distance  dependencies information introduced  by  self-attention. 
    Here, bi-directional LSTM encoders capture relationships among the words of the input text. 
    A multi-hop self-attention mechanism is applied to the resulting hidden states to produce a fixed sized embedding. 
    The different attention hops are constrained so as to reduce redundancy, ensuring that various semantic aspects of the input are included in the resulting embedding. 
    We refer the reader to the original paper for details.

  \subsection{Generation}
  Generation methods treat equation templates as strings of formal symbols.
  The production of a template is considered a sequence prediction problem conditioned on the word problem text. 
  By treating templates as sequences rather than monolithic structures, generation methods have the potential to learn finer-grained relationships between the input text and output template. 
  They also are the only methods studied here which can induce templates during inference which were not seen at training. 
  
  We generate equation templates with seq2seq models \cite{sutskever2014sequence} with attention mechanisms \cite{luong2015effective}.
  These models condition the token-by-token generation of the equation template on encodings of the word problem text. 
  Following~\citet{wang2017deep} we evaluate a seq2seq with LSTMs as the encoder and decoder.
  We also evaluate the use of Convolutional Neural Networks (CNNs) in the encoder and decoder.

\section{Experiments}
\subsection{Experimental Setup}
  \begin{table}[t]
  \centering
  \begin{center}
  \begin{tabular}{l|ccc}
  \hline  Dataset &  \# Quest. &  \# Templates &  \# Sent. \\  \hline

  {\sc Draw} & 1000 & 232 & 2.3k \\
  {\sc Mawps} & 2373 & 317 & 6.3k \\
  Math23K & 23164 & 3296 & 70.1k \\
  \hline
  \end{tabular}
  \end{center}
  \caption{\label{data-table} Dataset statistics. }
  \end{table}
  \paragraph{Datasets} 
  For comparison, we report solution accuracy on the Chinese language Math23K dataset \cite{wang2017deep},  and the English language {\sc Draw} \cite{upadhyay2015draw} and {\sc Mawps} \cite{koncel2016mawps} datasets. 
  Math23K and {\sc Mawps} consist of single equation problems, and {\sc Draw} contains both single and simultaneous equation problems. 
  Details on the datasets are shown in Table~\ref{data-table}.
  
  The Math23K dataset contains problems with possibly irrelevant quantities.
  To prune these quantities, we implement a significant number identifier (SNI) as discussed in \citet{wang2017deep}. 
    Our best accuracy for SNI is 97\%, slightly weaker than previous results.
  
  \paragraph{Implementation details}
    Our BiLSTM model's parameters are tuned on a validation set for each dataset.
     We also explore two modifications of the BiLSTM's embedding matrix $\bf{W_E}$, either by using pretrained GloVe embeddings \cite{pennington2014glove} or using the ELMo technique of \cite{peters2018deep} as implemented in the AllenNLP toolkit \cite{gardnerallennlp} with pretrained character embeddings.
     For seq2seq modeling, we use OpenNMT \cite{klein2017opennmt} with 500 dimensional hidden states and embeddings and a dropout rate of 0.3. 
     The CNN uses a kernel width of 3. 
     Optimization is done using SGD with a learning rate of 1, decayed by half if the validation perplexity does not decrease after an epoch.

\subsection{Results}
\begin{small}
\begin{savenotes}
  \begin{table}[t]
  \begin{center}
  \begin{tabular}{lccccc}
  \hline &  {\footnotesize DRAW} &  {\footnotesize MAWPS} &  {\footnotesize Math23K} \\
  \hline
  Oracle  & 79.0 & 84.8 & 87.0\\
  \hline
  \multicolumn{4}{c}{  Retrieval} \\ 
  \ \ Jaccard & 43.5 & 45.6 & 47.2 \\
  \ \ Cosine  & 29.5 & 38.8 & 23.8 \\ \hline
  \multicolumn{4}{c}{  Generation} \\ 
  \ \ LSTM & 15.0 & 25.6 & 51.96 \\
  \ \ CNN & 29.5 & 44.0 & 42.31 \\ \hline
  \multicolumn{4}{c}{  Classification} \\ 
  \ \ BiLSTM & 53.0 &  \bf 62.8 & 57.9 \\
  \ \ Self-Attention & \bf 53.5 & 60.4 & 56.8  \\ \hline
  State of the art & 52.0 & -- & \bf 64.7 \\ \hline
  \end{tabular}
  \end{center}
  \caption{\label{results-table} Accuracy of data-driven models for solving algebra word problems across 3 datasets.}
  \end{table}
\end{savenotes}
 \end{small}

  \begin{table}[t]
    \centering
    \begin{tabular}{lcc}
\hline &  {\footnotesize DRAW} &  {\footnotesize MAWPS} \\ \hline
    Classification & 53.0 & 62.8\\
    \ \ \ \ \ + GloVe & 42.0 & 31.6 \\ 
    \ \ \ \ \ + ELMo & 45.5 & 57.2 \\ \hline
    \end{tabular}
    \caption{Results of including pretrained word embeddings (GloVe) or character embeddings with learned layer weights (ELMo) into classification system.}
    \label{tab:abla}
\end{table}

  \begin{table*}[t]
  \begin{center}
  \begin{tabular}{p{0.25\columnwidth}|p{.8\textwidth}}
  \hline 
  \bf Type & \bf Problem Text \\
  \hline
 Semantic Limitations (36\%)& Kendra made punch for her friend's birthday party. She used 3/4 of a gallon of grape juice, 1/4 of a gallon of cranberry juice , and 3/5 of a gallon of club soda. How many gallons of punch did Kendra make? \\\cline{2-2}
  & Sandy went to the mall to buy clothes. She spent \$20 on shorts, \$10 on a shirt, and \$35 on a jacket. How much money did Sandy spend on clothes?
\\
  \hline \hline
  
  World Knowledge (19\%) & Mary began walking home from school, heading south at a rate of 3 miles per hour. Sharon left school at the same time heading north at 5 miles per hour. How long will it take for them to be 20 miles apart? \\
  \cline{2-2}
& If you purchase a membership for 100 dollars to receive 5\% off purchases, how much would you need to spend to pay off the membership?
\\
  \hline
  \end{tabular}
  \end{center}
  \caption{\label{errors-table} Example Error Categories and Occurrence Rates.}
  \end{table*}
  Table \ref{results-table} reports the accuracies of the data-driven models for solving algebra word problems.
  The classification models perform better than retrieval or generation models, despite their limited modeling power.
  The self-attention classification model performs well across all datasets.  
  For the largest dataset (Math23K), a simple, well-tuned classifier can outperform the more sophisticated sequence-to-sequence and self-attention models.
  
  Table~\ref{tab:abla} shows results of augmenting the classifier with pretrained word and character embeddings. 
  Neither of these methods help over the English language data. 
  It appears that the ELMo technique may require more training examples before it can improve solution accuracy. 
  
  The previous state of the art model for the {\sc Draw} dataset is described in \citet{upadhyay2015draw}. 
  The state of the art for Math23K, described in \citet{wang2017deep}, uses a hybrid Jaccard retrieval and seq2seq model. 
  All models shown here fall well short of the highest possible classification/retrieval accuracy, shown in Table~\ref{results-table} as ``Oracle''. 
  This gap invites a more detailed error analysis regarding the possible limitations of data-driven solvers.

\subsection{Error Analysis}

    Despite the sophistication of these data-driven models, they still do not achieve optimal performance. 
    A closer analysis of the errors these models make can illuminate the reason for this gap.
    
    Consider Table \ref{errors-table}, which illustrates two classes of errors made by data-driven systems. 
    Both stem from incomplete knowledge on the part of the learning algorithm.
    But it is worth distinguishing the ``semantic limitations'' errors as this kind of information (subset relations, counts of non-numerical entities) may be possible to extract from the data provided, given a sufficiently powerful modeling technique.
    
    The second class of errors, labeled ``world knowledge", are impossible to extract from the math data alone.
    Consider the first example of people walking in different directions.
    To solve this problem, it is necessary to know that ``north" and ``south" are away from each other.
    Complicating the problem, suppose Sharon walked east instead of north.
    Then the relationship between east and south would impact the problem semantics. 
    This kind of knowledge is beyond what is conveyed in any dataset of math word problems, and is a known problem for many NLP applications. 
   
\section{Related Work}

Semantic solvers provide some scaffolding for the grounding of word problem texts to equations.
\citet{mitralearning} solve simple word problems by categorizing their operations as part-whole, change, or comparison.
\citet{shi15automatically} learn a semantic parser by semi-automatically inducing 9600 grammar rules over a dataset of number word problems. 
Works such has \citet{roy2015solving} and \citet{koncel2015parsing} treat arithmetic word problem templates as equation trees and perform efficient tree-search by learning how to combine quantities using textual information.
\citet{roy2017unit} advance this approach by considering unit consistency in the tree-search procedure. 
\citet{wang2018mathdqn} advance this line of work even further by modeling the search using deep Q-learning. 
Still, these semantic approaches are limited by their inability to model systems of equations as well as use of hand-engineered features. 

Data-driven math word problem solvers include \citet{kushman2014learning}, who learn to predict equation templates and subsequently align numbers and unknowns from the text. 
\citet{zhou2015learn} only assign numbers to the predicted template, reducing the search space significantly. 
More recently, \citet{wang2017deep} provide a large dataset of Chinese algebra word problems and learn a hybrid model consisting of both retrieval and seq2seq components. 
The current work extends these approaches by exploring advanced techniques in data-driven solving.

    \section{Conclusion}
    We have thoroughly examined data-driven models for automatically solving algebra word problems, including retrieval, classification, and generation techniques.
    We find that a well-tuned classifier outperforms generation and retrieval on several datasets. 
    One avenue for improving performance is to ensemble different models. 
    However, in light of the error analysis provided, the incorporation of semantic and world knowledge will be necessary to achieve maximal success.

\bibliographystyle{acl_natbib}
\bibliography{references}

\appendix

\end{document}